\documentclass{article}

\usepackage[preprint]{neurips_2025}

\usepackage[utf8]{inputenc} 
\usepackage[T1]{fontenc}    
\usepackage{hyperref}       
\usepackage{url}            
\usepackage{booktabs}       
\usepackage{amsfonts}       
\usepackage{nicefrac}       
\usepackage{microtype}      
\usepackage{xcolor}         

\usepackage{subcaption}
\usepackage{graphicx}
\usepackage{wrapfig}
\usepackage{amsmath}
\usepackage{enumitem}
\usepackage[linesnumbered,ruled,vlined]{algorithm2e}
\usepackage{multirow}
\usepackage{bbding}
\usepackage{dsfont}

\newtheorem{theorem}{Theorem}

\title{Instance-Prototype Affinity Learning for Non-Exemplar Continual Graph Learning}

\author{%
  \textbf{Lei Song\textsuperscript{1}, Jiaxing Li\textsuperscript{1}, Shihan Guan\textsuperscript{1}, Youyong Kong\textsuperscript{1,2}\thanks{Corresponding Author.} }\\
  \textsuperscript{1}Jiangsu Provincial Joint International Research Laboratory of Medical Information Processing,\\School of Computer Science and Engineering, Southeast University\\
  \textsuperscript{2}Key Laboratory of New Generation Artificial Intelligence Technology and Its Interdisciplinary\\Applications (Southeast University), Ministry of Education, China\\
  \texttt{\{230238577, jiaxing\_li, 230228507, kongyouyong\}@seu.edu.cn}
}

\begin{document}

\maketitle

\begin{abstract}
Graph Neural Networks (GNN) endure catastrophic forgetting, undermining their capacity to preserve previously acquired knowledge amid the assimilation of novel information. Rehearsal-based techniques revisit historical examples, adopted as a principal strategy to alleviate this phenomenon. However, memory explosion and privacy infringements impose significant constraints on their utility. Non-Exemplar methods circumvent the prior issues through Prototype Replay (PR), yet feature drift presents new challenges. In this paper, our empirical findings reveal that Prototype Contrastive Learning (PCL) exhibits less pronounced drift than conventional PR. Drawing upon PCL, we propose Instance-Prototype Affinity Learning (IPAL), a novel paradigm for Non-Exemplar Continual Graph Learning (NECGL). Exploiting graph structural information, we formulate Topology-Integrated Gaussian Prototypes (TIGP), guiding feature distributions towards high-impact nodes to augment the model's capacity for assimilating new knowledge. Instance-Prototype Affinity Distillation (IPAD) safeguards task memory by regularizing discontinuities in class relationships. Moreover, we embed a Decision Boundary Perception (DBP) mechanism within PCL, fostering greater inter-class discriminability. Evaluations on four node classification benchmark datasets demonstrate that our method outperforms existing state-of-the-art methods, achieving a better trade-off between plasticity and stability. 
\end{abstract}

\section{Introduction}
\label{Introduction}

As a potent paradigm for graph data analysis, Graph Neural Networks (GNN)~\cite{hamilton2017inductive, kipf2016semi, velivckovic2017graph, xu2018powerful} have garnered significant academic attention in recent years. However, most existing studies~\cite{bi2023mm, dong2022protognn, wu2019simplifying} adhere to a static data regime, where the complete training set is available upfront and model parameters remain immutable after initial optimization. The real world exhibits an intrinsically dynamic nature, with information incessantly generated, such as user interactions on social media or the dissemination of domain-specific publications, posing considerable exigencies for static modeling paradigms. Continual Graph Learning (CGL) endeavors to assimilate novel knowledge while retaining previously acquired representations. Nonetheless, distributional disparities across tasks frequently precipitate catastrophic forgetting, manifesting as marked performance degradation on earlier tasks. Rehearsal-based approaches~\cite{arani2022learning, zhou2021overcoming} ameliorate this challenge by retrospectively incorporating a curated subset of exemplars from prior tasks. However, since representative exemplars are retained in the memory buffer for each task, the memory footprint escalates with longer task sequences, potentially culminating in memory explosion. Moreover, in privacy-sensitive contexts, access to raw examples may be constrained.

To surmount the challenges identified above, we examine Non-Exemplar Continual Graph Learning (NECGL), a more stringent paradigm that prohibits access to prior raw examples when encountering new tasks. In this context, catastrophic forgetting is further exacerbated. Existing Non-Exemplar approaches~\cite{li2024fcs, magistri2024elastic, ren2023incremental, wang2023non} revisit historical prototype representations---the embeddings of prior classes in the encoder's latent space---to overcome this limitation via Prototype Replay (PR). However, successive updates to model parameters render earlier class prototypes progressively obsolete, giving rise to feature drift---reflected in their misalignment within the evolving feature space. Most existing approaches rectify prior class prototypes through feature drift compensation techniques~\cite{gomez2024exemplar, yu2020semantic} post-training, or apply knowledge distillation~\cite{hinton2015distilling, zhu2021prototype} online to regulate the evolution of feature space. While feature drift remains an inherent challenge in NECGL, empirical evidence suggests that Prototype Contrastive Learning (PCL)~\cite{li2025inductive} induces less drift than conventional PR.
\begin{wrapfigure}{r}{0.6\textwidth}
    \vspace{-10pt}
    \centering
    \begin{minipage}[b]{0.5\linewidth}
        \centering
        \includegraphics[width=\linewidth, trim=13 20 20 20, clip]{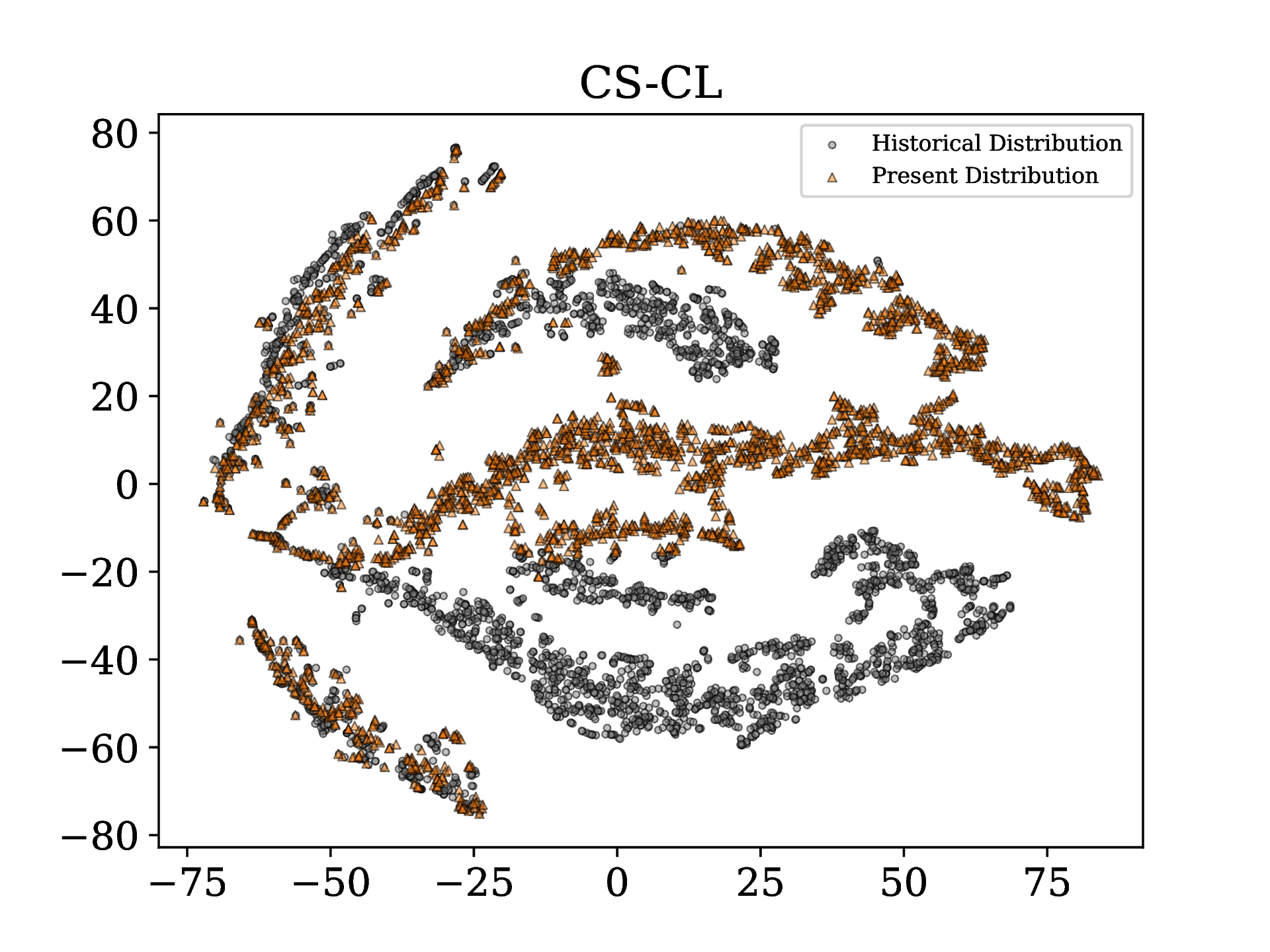}
    \end{minipage}
    \hspace{-10pt}
    \begin{minipage}[b]{0.5\linewidth}
        \centering
        \includegraphics[width=\linewidth, trim=13 20 20 20, clip]{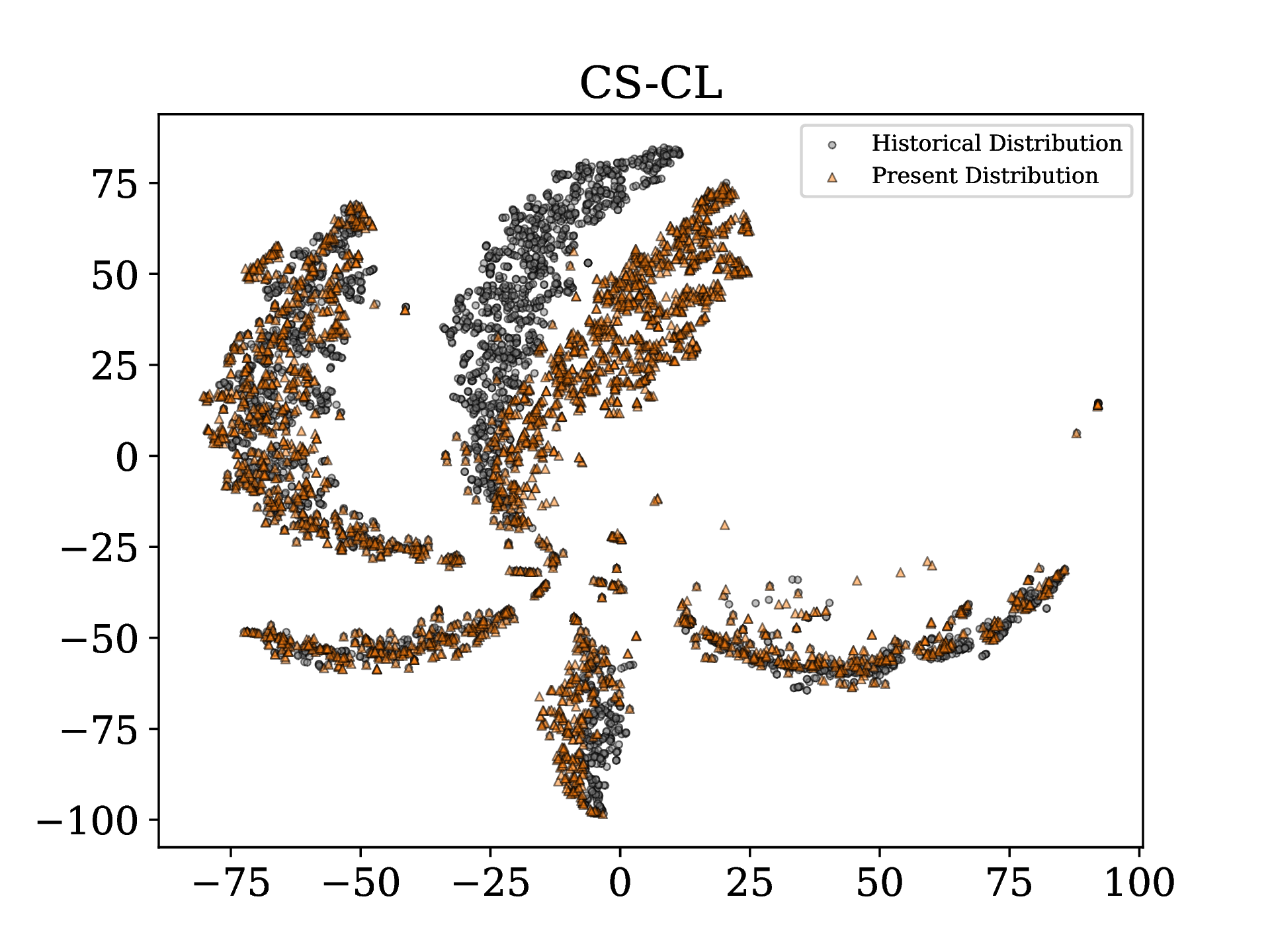}
    \end{minipage}
    
    \vspace{2pt}
    
    \begin{minipage}[b]{0.5\linewidth}
        \centering
        \includegraphics[width=\linewidth, trim=13 20 20 20, clip]{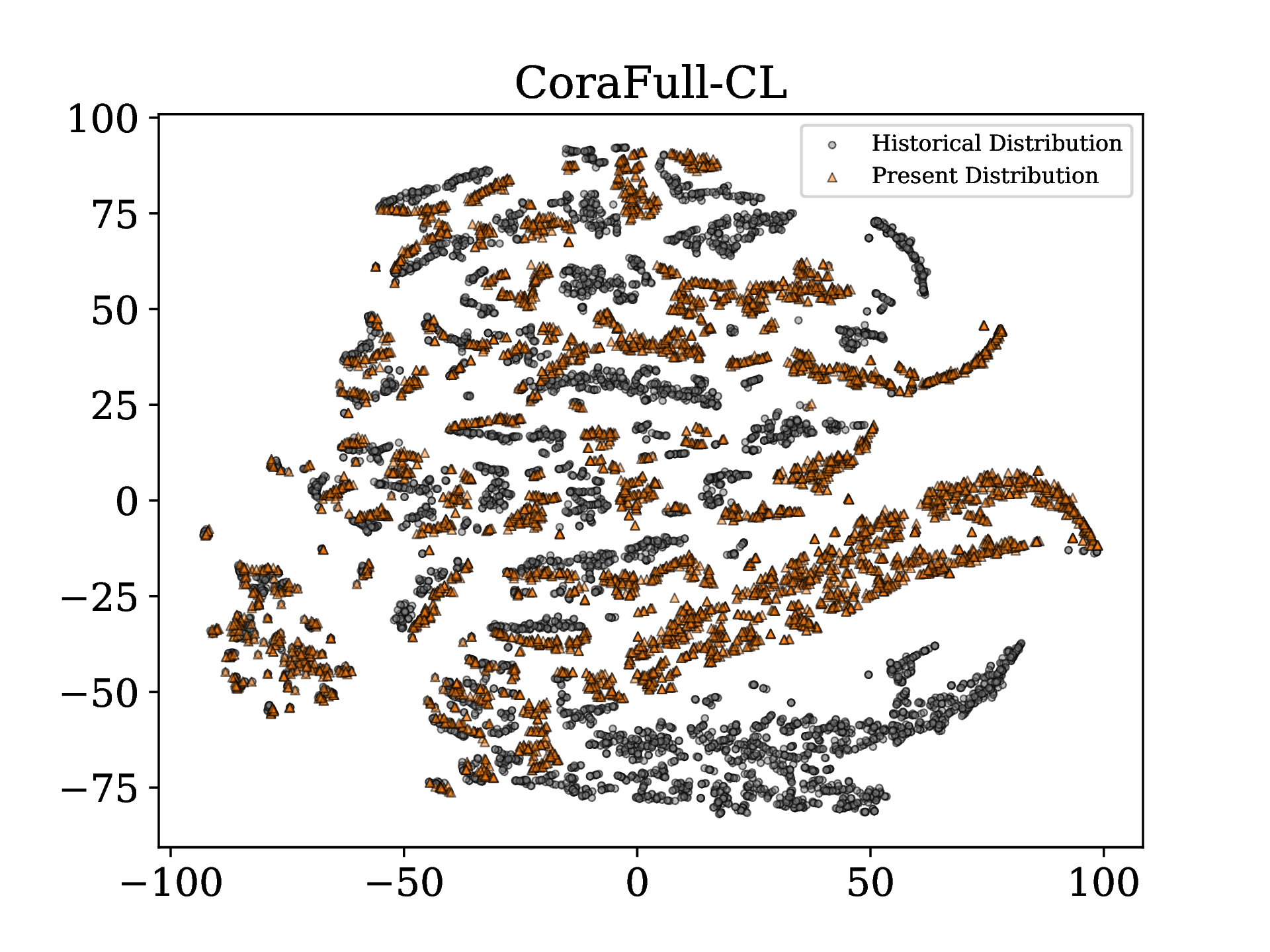}
    \end{minipage}
    \hspace{-10pt}
    \begin{minipage}[b]{0.5\linewidth}
        \centering
        \includegraphics[width=\linewidth, trim=13 20 20 20, clip]{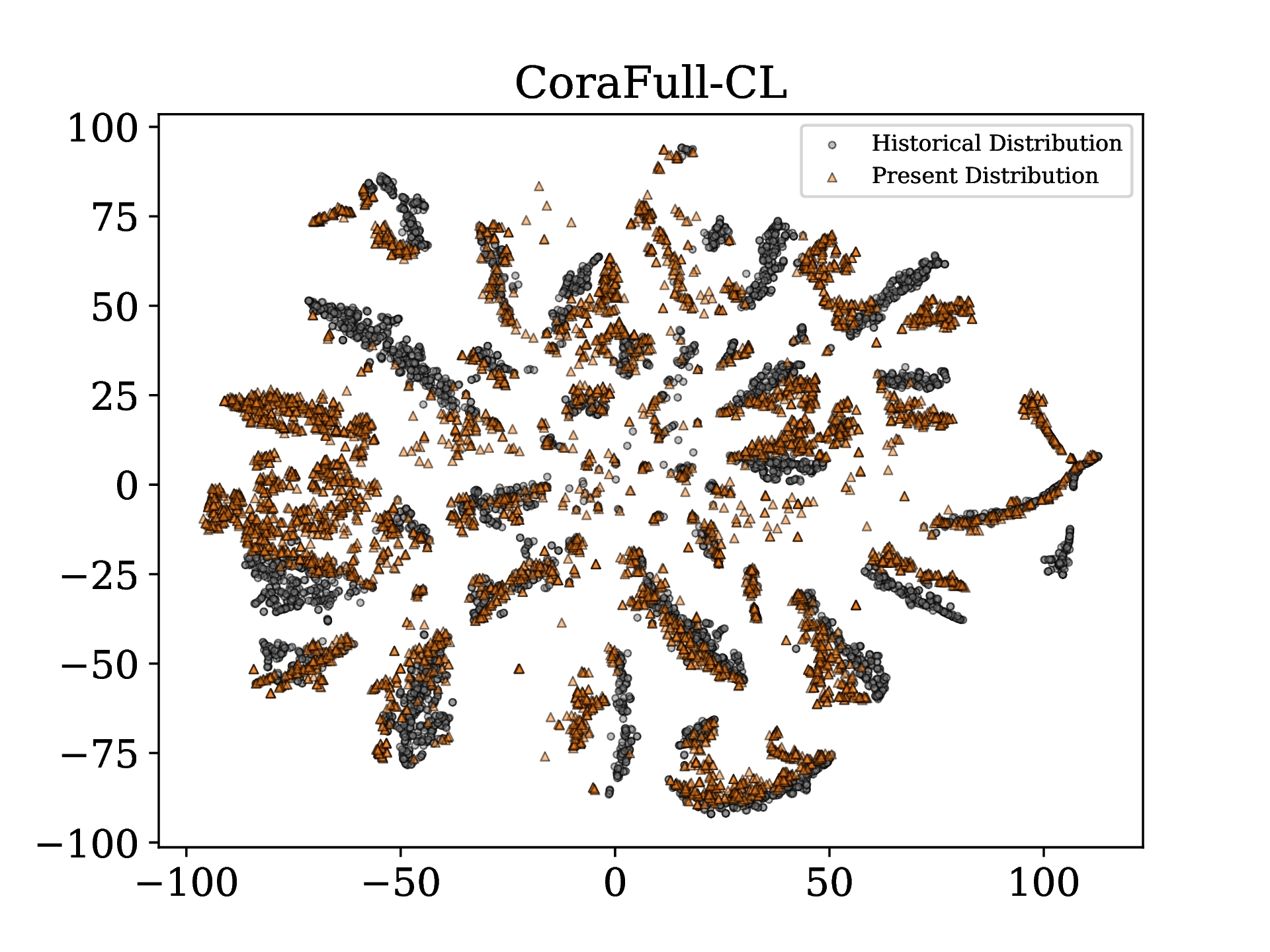}
    \end{minipage}
    \caption{Visualization of feature drift for conventional Prototype Replay (left column) and Prototype Contrastive Learning (right column) on the CS-CL and CoraFull-CL datasets.}
    \label{fig1}
    \vspace{-10pt}
\end{wrapfigure}
As shown in Figure~\ref{fig1}, we visualize the extent of feature drift on base task samples across the CS-CL and CoraFull-CL datasets. Conventional PR formulates training with a cross-entropy objective, drawing on prior class prototypes for classification. As model parameters evolve, prototype misalignment intensifies, resulting in greater prediction errors and exacerbated feature drift. PCL encourages learning relational structures between instances and prototypes, explicitly distinguishing previously encountered classes from novel ones, thus attenuating task interference. To elucidate this phenomenon, we present Theorem~\ref{theo1} in Appendix~\ref{appendix:a}, along with a rigorous theoretical proof. In contrast to Graph Contrastive Learning~\cite{hassani2020contrastive, xu2021infogcl}, PCL inherently establishes positive pairs between instances and class-aligned prototypes, and negative pairs with inter-class prototypes, obviating the need for predefined augmentation strategies such as edge perturbation or attribute masking. Motivated by this observation, we propose a novel NECGL paradigm built upon PCL.

While existing Non-Exemplar methods sustain model stability through PR and knowledge distillation, notable limitations persist. First, most existing works~\cite{cheng2024efficient, li2024fcs, li2025inductive, magistri2024elastic, ren2023incremental} derive prototypes by averaging feature representations, yet this isotropic way neglects the varying importance of nodes. For graph-structured data, node significance is shaped by unique topologies and neighbor influences. Second, feature distillation is extensively applied in Non-Exemplar methods due to its inherent plug-and-play functionality. Nonetheless, recent research~\cite{magistri2024elastic} indicates that it imposes excessive constraints on the feature space, inhibiting model plasticity. Third, a solitary prototype is often insufficient to capture the complete distribution of a class. In the context of PCL, the exclusive reliance on a single class prototype can give rise to inter-class ambiguity, thereby exacerbating catastrophic forgetting.

To overcome the above limitations, we propose Instance-Prototype Affinity Learning (IPAL), a novel framework tailored for NECGL. We evaluate node impact via the PageRank algorithm~\cite{page1999pagerank} and generate Topology-Integrated Gaussian Prototypes (TIGP), directing class distributions towards high-impact nodes to facilitate the assimilation of new knowledge. To combat catastrophic forgetting, we propose Instance-Prototype Affinity Distillation (IPAD), aligning instance-prototype relationships for more flexible regularization of the feature space. Notably, IPAD seamlessly integrates with PCL, providing distinct advantages over feature distillation. Moreover, we embed Decision Boundary Perception (DBP) mechanism into PCL to promote sharper inter-class delineation by repelling instances proximal to decision boundaries.

\textbf{Contributions.} The main contributions of this paper are as follows: \romannumeral1) We propose IPAL, a novel paradigm tailored for NECGL that strikes a favorable trade-off between stability and plasticity; \romannumeral2) We utilize the PageRank algorithm to generate more robust TIGP, integrating graph topology into the prototype computation to amplify learning capacity; \romannumeral3) We design IPAD, a knowledge distillation method inherently compatible with PCL, enabling more flexible retention of prior knowledge; \romannumeral4) We incorporate the DBP mechanism into the PCL objective for clearer inter-class separation; \romannumeral5) Extensive experiments on four node classification benchmark datasets demonstrate that our proposed IPAL outperforms existing state-of-the-art methods in the Non-Exemplar scenario.

\section{Related Work}
\label{Related_Work}

\textbf{Continual Graph Learning.} CGL seeks to assimilate new knowledge while preventing GNN from forgetting historical knowledge. Prior studies adopted regularization, rehearsal, parameter isolation, or their combinations to mitigate catastrophic forgetting. Regularization-based methods~\cite{aljundi2018memory, kirkpatrick2017overcoming, li2017learning, liu2021overcoming} reinforce constraints on pivotal parameters by quantifying their significance, or facilitate output alignment via knowledge distillation applied to the model's logits. Rehearsal-based methods~\cite{liu2023cat, zhang2022sparsified, zhou2021overcoming} store task-specific exemplars in a memory buffer for replay when learning new tasks. Parameter isolation methods~\cite{niu2024replay, zhang2023continual} prevent inter-task interference by assigning separate parameters or learning task-specific submodules. This work focuses on rehearsal-based methods for their superior performance and closer resemblance to human learning. Despite their efficacy, prolonged task sequences can impose substantial memory burdens, and privacy restrictions may limit access to historical data.

\textbf{Non-Exemplar Continual Learning.} Non-Exemplar Continual Learning (NECL) updates models without revisiting prior raw examples. Existing studies~\cite{li2024fcs, magistri2024elastic, ren2023incremental, wang2023non} circumvent memory and privacy concerns via PR, rather than replaying raw examples. However, feature drift remains a fundamental flaw of these approaches. While feature distillation curbs substantial variations in the feature space, \cite{magistri2024elastic} argued it induces excessive regularization, yielding performance akin to freezing the backbone after the base task~\cite{petit2023fetril}. To reconcile historical prototypes with the new feature space, \cite{li2025inductive, magistri2024elastic, wang2023non, yu2020semantic, zhai2024fine} quantified prototype drift via statistical measures derived from new task samples, while \cite{cheng2024efficient, gomez2024exemplar, li2024fcs} harnessed learnable neural networks for adaptive compensation. In this paper, we empirically observe from Figure~\ref{fig1} that PCL exhibits less prototype drift than conventional PR trained with cross-entropy, which motivates the proposal of IPAL.

\section{Preliminaries}
\label{Preliminaries}

\textbf{Problem Formulation.} This paper explores the Class-Incremental Learning (CIL) setting, wherein a GNN is optimized consecutively over a task sequence $\mathcal{T}=\{\mathcal{T}_{0}, \mathcal{T}_{1},...,\mathcal{T}_{N}\}$ with $\left|\mathcal{T}\right|=N+1$. Each task $\mathcal{T}_{t\le N}=\{\mathcal{G}_{t}, \mathcal{Y}_{t}\}$ is defined as a semi-supervised node classification task, where $\mathcal{G}_{t}=\{\mathcal{V}_{t}, \mathcal{E}_{t}\}$ denotes the task graph with node set $\mathcal{V}_{t}$ and edge set $\mathcal{E}_{t}$. $\mathcal{E}_{t}$ can be expressed through a binary adjacency matrix $\mathbf{A}_{t}$, where 1 indicates an edge and 0 its absence. The label set is given by $\mathcal{Y}_{t}=\{y^{1}_{t}, y^{2}_{t},..., y^{c_{t}}_{t}\}$, with $\mathcal{Y}_{i}\cap \mathcal{Y}_{j}=\emptyset$ for $i\neq j$. In the NECGL paradigm, the base task $\mathcal{T}_{0}$ typically comprises considerably more data than each incremental task $\mathcal{T}_{t}$ ($t>0$), facilitating GNN pretraining for improved incremental adaptation, i.e., $c_{0}\gg c_{t}$. Crucially, when learning new tasks, access to data from prior tasks is rigorously restricted, allowing only the current task's data. Our aim is to train a GNN on the task sequence $\mathcal{T}$ to attain superior performance across all previously encountered tasks. For a model with an encoder $\mathcal{F}_{\theta_t}(\cdot)$  and a linear classifier $g_{\phi_t}(\cdot)$, existing approaches predominantly employ PR to alleviate catastrophic forgetting, with the optimization objective formalized as follows:
\begin{equation}\label{eq1}
\mathcal{L}_{PR}=\mathbb{E}_{(x_t, y_t)\in \mathcal{T}_{t}}\left[-y_{t}^c\log g_{\phi_t}(\mathcal{F}_{\theta_t}(x_t^c))\right] + \mathbb{E}_{(f_\mathcal{M}^m, y_\mathcal{M}^m)\in \mathcal{M}}\left[-y_\mathcal{M}^{m}\log g_{\phi_t}(f_\mathcal{M}^m)\right],
\end{equation}
where $\mathcal{M}$ denotes the memory buffer storing historical class prototypes (i.e., Gaussian distributions with mean $\mu_m$ and diagonal covariance $\sigma_m^{2}$), from which feature representations $f_\mathcal{M}^{m}$ reflecting past task distributions are sampled for replay during new task learning.

\textbf{Prototype Contrastive Learning.} PCL, rooted in Prototypical Networks~\cite{snell2017prototypical} from few-shot learning, captures the semantic association between instance-wise and class-wise representations. The optimization objective is formally stated as follows:
\begin{equation}\label{eq2}
\mathcal{L}_{PCL}=\mathbb{E}_{(x_t, y_t)\in \mathcal{T}_{t}}\left[-\log\frac{e^{\mathcal{F}_{\theta_t}(x_t^c)^{\top}\cdot \mu_c/\tau}}{e^{\mathcal{F}_{\theta_t}(x_t^c)^{\top}\cdot \mu_c/\tau}+\sum_{j\neq c}e^{\mathcal{F}_{\theta_t}(x_t^c)^{\top}\cdot \mu_j/\tau}}\right],
\end{equation}
where $\mu_{k\in \mathcal{Y}_{0:t}}=\frac{\sum_{(x,y)\in \mathcal{T}}\mathds{1}\{y=k\}\mathcal{F}_\theta(x)}{\sum_{(x,y)\in \mathcal{T}}\mathds{1}\{y=k\}}$ denotes the prototype for the class $k$, and $\tau$ is the temperature scaling factor that controls the distributional smoothness. In this paper, offline prototypes from prior tasks in $\mathcal{M}$ and current task instances form negative pairs, while online prototypes and label-matching instances from the current task form positive pairs.

\section{Methodology}
\label{Methodology}

\begin{figure}[t]
    \centering
    \includegraphics[width=\linewidth, trim=28 18 115 203, clip]{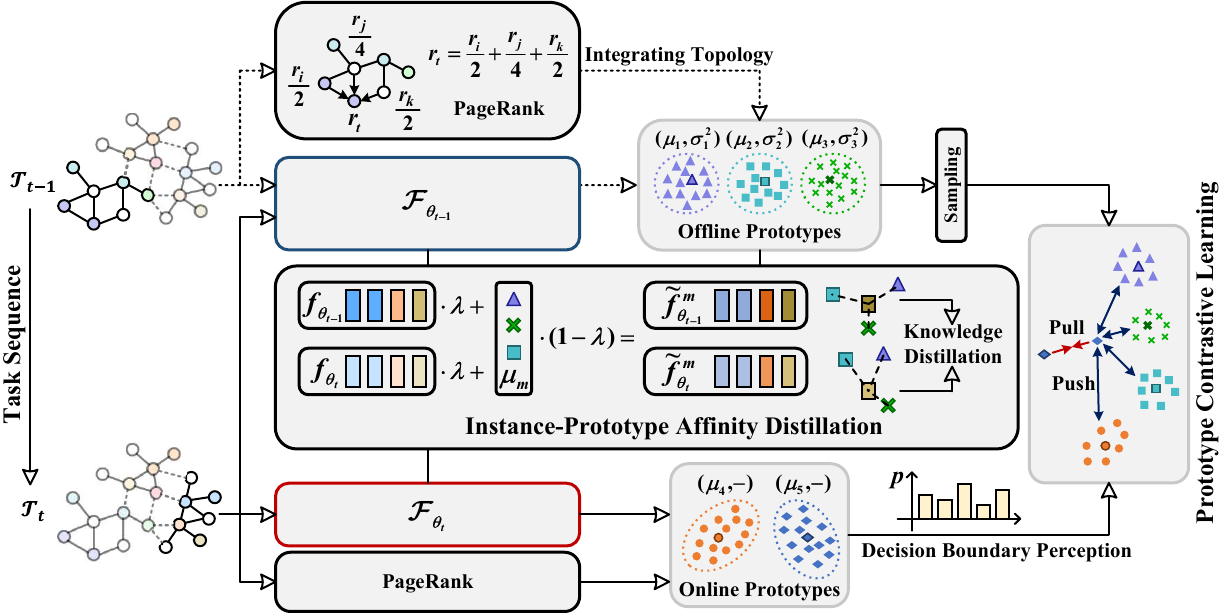}
    \caption{The overall pipeline of the proposed IPAL framework. Upon the culmination of task $\mathcal{T}_{t-1}$, the TIGP are derived and offline archived in memory buffer $\mathcal{M}$. Following the onset of task $\mathcal{T}_{t}$, online prototypes are dynamically updated and integrated with offline prototypes for PCL. In this regard, IPAD safeguards prior task memory via relational distillation, while DBP ensures the clear demarcation of newly encountered classes. Best viewed in color.}
    \label{fig2}
\end{figure}

Figure \ref{fig2} outlines the overall pipeline of the proposed IPAL. In this section, we provide a detailed analysis of each component.

\subsection{Topology-Integrated Gaussian Prototypes}

NECGL prohibits access to raw examples from prior tasks. To combat the intensified catastrophic forgetting, existing Non-Exemplar methods model each encountered class $k$ with a Gaussian distribution $\mathcal{N}(\mu_k, \sigma_{k}^{2})$ after training, and retain it as a class prototype in the memory buffer $\mathcal{M}$. Although somewhat effective, treating all nodes uniformly in graph-structured data is untenable. Owing to the distinctiveness of graph topology, nodes within disparate neighborhoods exert differential influence on their adjacent counterparts. For example, prominent celebrities often possess a vast following, and their actions tend to wield greater influence on society. Indeed, such nodes tend to be more indicative than low-degree ones. We steer the model to align class distributions with high-impact nodes, aiding the assimilation of new knowledge. Drawing inspiration from \cite{page1999pagerank}, node importance is evaluated via the PageRank algorithm, formalized as follows:
\begin{equation}
    r_t=\alpha\sum_{j\in \mathcal{N}^{in}(t)}\frac{r_j}{d_{j}^{+}}+(1-\alpha),
\end{equation}
where $\mathcal{N}^{in}(t)$ denotes the set of incoming neighbors of node $t$, and $d_{j}^{+}=\left|\mathcal{N}^{out}(j)\right|$ represents the out-degree of node $j$. $r_t$ is the PageRank for node $t$, and $\alpha$ is the damping factor. We then reweight the node contributions with the aid of PageRank to compute the TIGP for each class as follows:
\begin{equation}\label{eq4}
    \mu_k=\frac{\sum_{(x,y)\in\mathcal{T}}r_x\mathds{1}\{y=k\}\mathcal{F}_\theta(x)}{\sum_{(x,y)\in\mathcal{T}}r_x\mathds{1}\{y=k\}},\quad\sigma^{2}_k=\mathrm{diag}\left(\frac{\sum_{(x,y)\in\mathcal{T}}r_x\mathds{1}\{y=k\}(\mathcal{F}_\theta(x)-\mu_k)^2}{\sum_{(x,y)\in\mathcal{T}}r_x\mathds{1}\{y=k\}}\right).
\end{equation}
Importantly, PageRank is computed once per task, avoiding recalculation in later iterations and imposing no extra burden on training. Moreover, offline prototypes are derived for all classes at the conclusion of each task, while dynamic online prototypes are instantiated for emerging classes throughout the PCL process to promote the integration of new knowledge. We visualize the performance heatmaps for mean-based prototypes and TIGP on three benchmark datasets. 
\begin{wrapfigure}{r}{0.5\textwidth} 
    \vspace{-12pt}
    \centering
    \begin{minipage}[b]{0.33\linewidth}
        \centering
        \includegraphics[width=\linewidth]{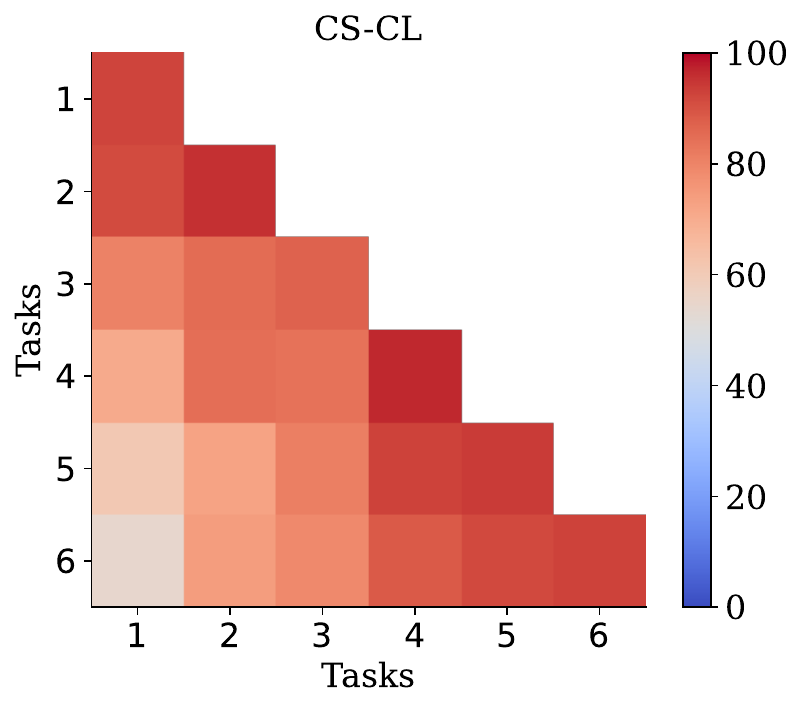}
    \end{minipage}
    \hspace{-5pt}
    \begin{minipage}[b]{0.33\linewidth}
        \centering
        \includegraphics[width=\linewidth]{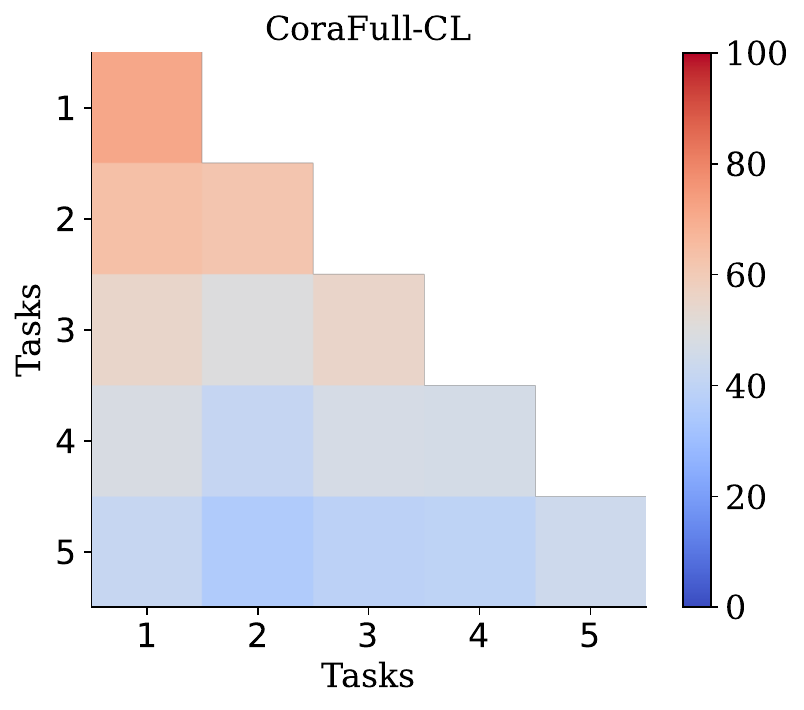}
    \end{minipage}
    \hspace{-5pt}
    \begin{minipage}[b]{0.33\linewidth}
        \centering
        \includegraphics[width=\linewidth]{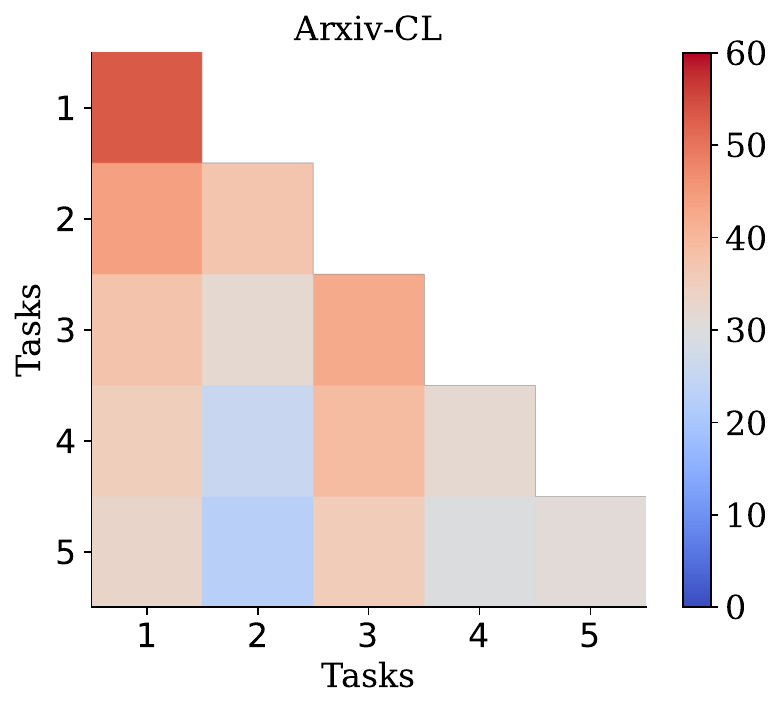}
    \end{minipage}

    \vspace{-3pt}
    
    \begin{minipage}[b]{0.33\linewidth}
        \centering
        \includegraphics[width=\linewidth]{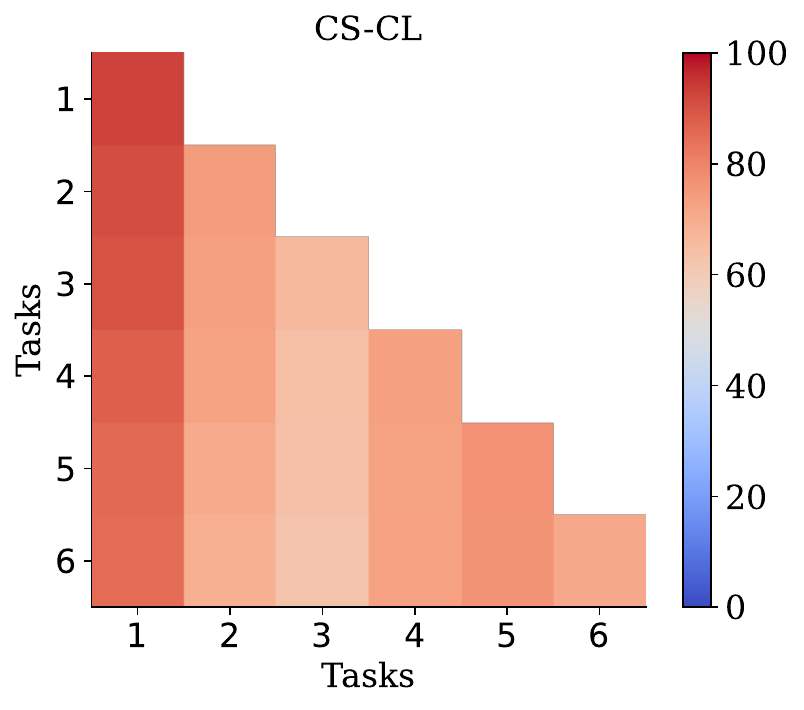}
    \end{minipage}
    \hspace{-5pt}
    \begin{minipage}[b]{0.33\linewidth}
        \centering
        \includegraphics[width=\linewidth]{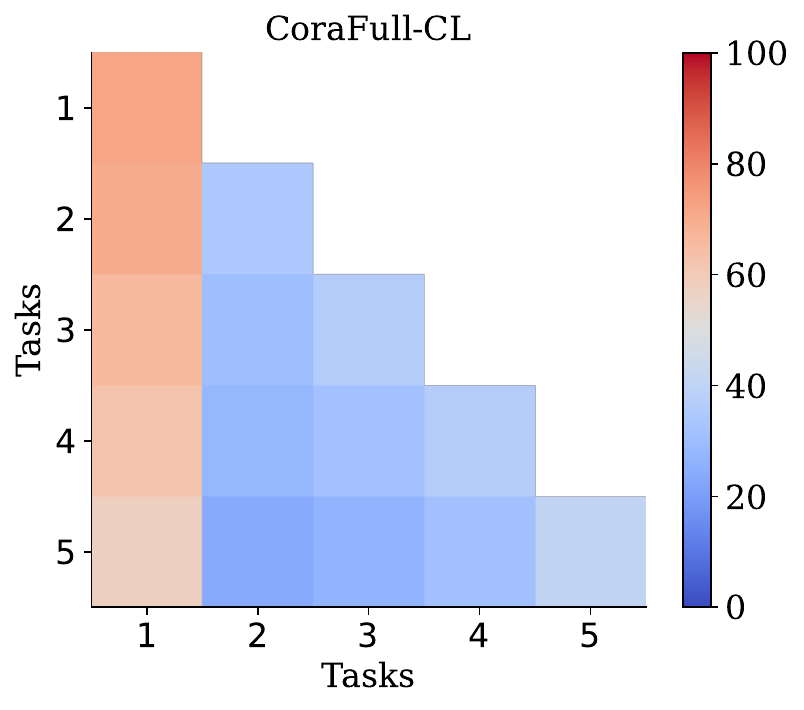}
    \end{minipage}
    \hspace{-5pt}
    \begin{minipage}[b]{0.33\linewidth}
        \centering
        \includegraphics[width=\linewidth]{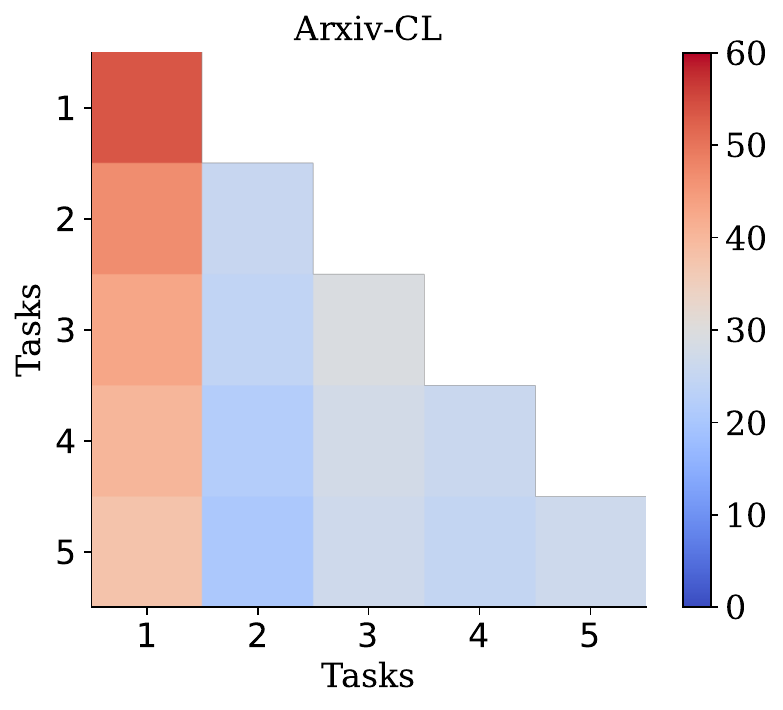}
    \end{minipage}
    
    \vspace{-3pt}
    
    \begin{minipage}[b]{0.33\linewidth}
        \centering
        \includegraphics[width=\linewidth]{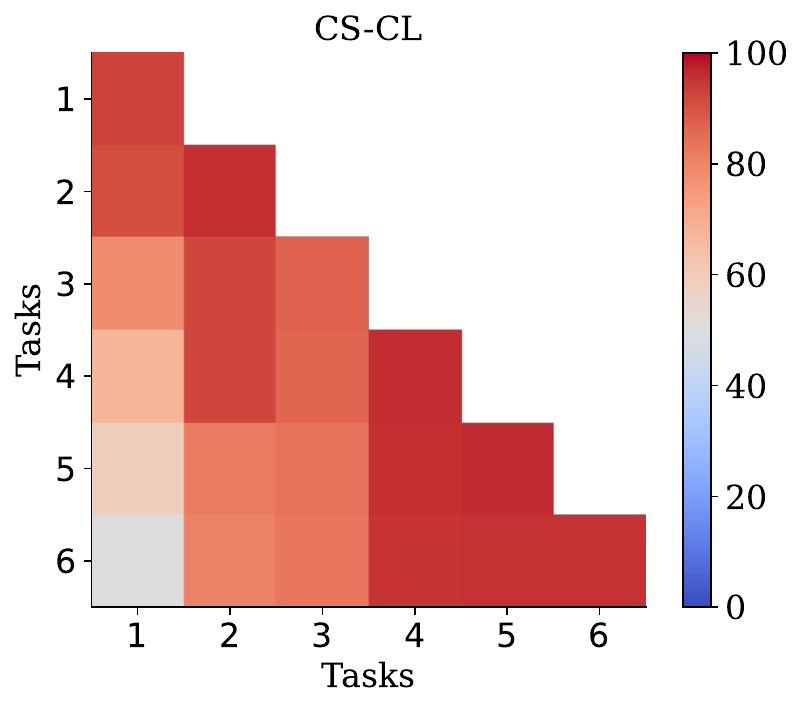}
    \end{minipage}
    \hspace{-5pt}
    \begin{minipage}[b]{0.33\linewidth}
        \centering
        \includegraphics[width=\linewidth]{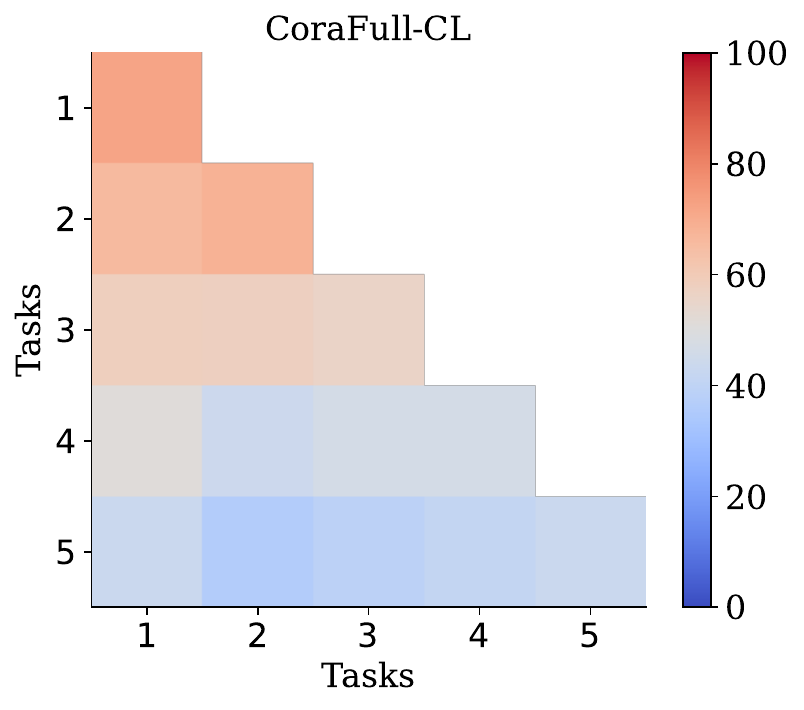}
    \end{minipage}
    \hspace{-5pt}
    \begin{minipage}[b]{0.33\linewidth}
        \centering
        \includegraphics[width=\linewidth]{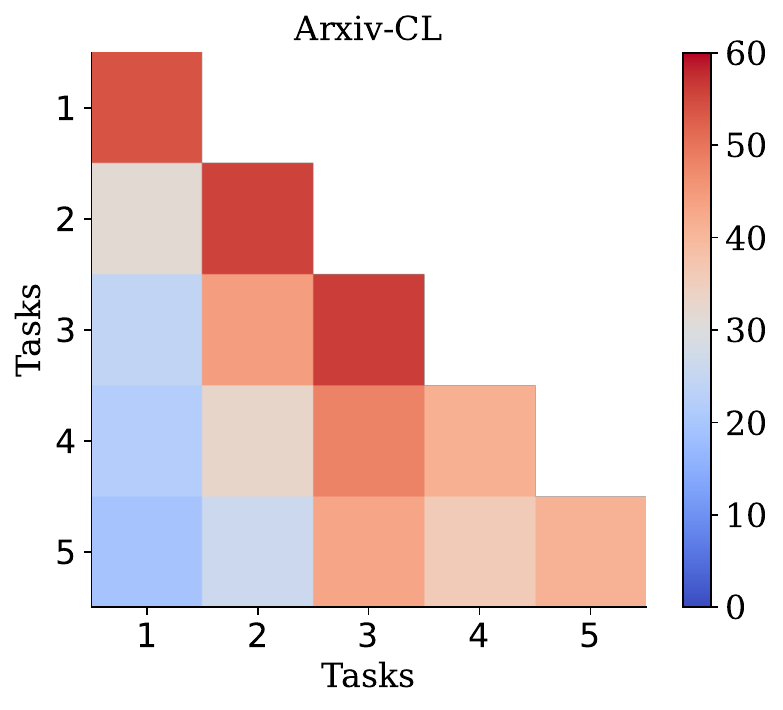}
    \end{minipage}
    \caption{Performance heatmaps on three datasets are shown. Top row: Mean-based prototypes for PCL; Middle row: Feature distillation to mitigate feature drift; Bottom row: Our proposed IPAL, which integrates TIGP and IPAD to better balance the trade-off between plasticity and stability.}
    \label{fig3}
    \vspace{-40pt}
\end{wrapfigure}
Figure \ref{fig3} demonstrates that TIGP markedly surpasses mean-based prototypes, yielding notably enhanced plasticity.

\subsection{Instance-Prototype Affinity Distillation}

A fundamental flaw of NECGL is feature drift (refer to Figure \ref{fig1}), wherein continual updates to the GNN render previously stored prototypes in memory buffer $\mathcal{M}$ increasingly obsolete and incompatible with the evolving feature space, thereby exacerbating the risk of catastrophic forgetting during replay. A direct yet effective strategy involves replaying historical prototypes to the classifier $g_{\phi_t}(\cdot)$, while regularizing the current parameters $\theta_t$ toward their previous optimum $\theta_{t-1}$. A prevalent paradigm in constraining $\theta_t$ is feature distillation, formally defined as follows:
\begin{equation}
    \mathcal{L}_{FD}=\mathbb{E}_{(x_t,y_t)\in\mathcal{T}_t}\|\mathcal{F}_{\theta_t}(x_t)-\mathcal{F}_{\theta_{t-1}}(x_t)\|_{2}.
\end{equation}
Only the data from the current task $\mathcal{T}_t$ is permissible for use. While it substantially alleviates the drift issue, we find that feature distillation may sacrifice model plasticity for greater stability. As depicted in the middle row of Figure~\ref{fig3}, feature distillation compels the model to predominantly preserve the initial feature space, with incremental learning merely embedding new class distributions within this confined space, thus impeding the assimilation of novel knowledge. Our observation concurs with \cite{magistri2024elastic}, which reveals that feature distillation may degrade continual learning into merely fine-tuning the classifier while freezing the backbone after the base task.

To circumvent the rigidity of feature distillation, we resort to IPAD, which upholds intra- and inter-class instance–prototype relations for more flexible regularization. Moreover, it exhibits intrinsic compatibility with the PCL objective function, enabling more reliable control. To elaborate, the embeddings $f_{\theta_{t-1}}$ and $f_{\theta_{t}}$ for the current task samples $(x_t,y_t)\in\mathcal{T}_t$ are derived from the preceding and current GNNs, $\mathcal{F}_{\theta_{t-1}}(\cdot)$ and $\mathcal{F}_{\theta_{t}}(\cdot)$, respectively. Next, we utilize the Mixup strategy~\cite{wang2021mixup, zhang2017mixup} to interpolate them with prior class prototypes, synthesizing virtual features that maintain close proximity to the prototypes, thus ensuring clear class delineation. The formalized expression is as follows:
\begin{equation}
    \tilde{f}_{\theta_{t-1}}^{m}=\lambda f_{\theta_{t-1}} + (1-\lambda)\mu_m,\quad \tilde{f}_{\theta_{t}}^{m}=\lambda f_{\theta_{t}} + (1-\lambda)\mu_m,
\end{equation}
where $\lambda\in[0,0.4]$ is drawn from a Beta distribution, i.e., $\lambda\sim\mathrm{Beta}(9,21)$. For efficiency, only a subset of $\mathcal{T}_t$ is engaged in Mixup. To prevent semantic noise in synthetic features, we further enforce pseudo-label filtration for label consistency:
\begin{equation}
    \mathbf{idx}=\underset{m\in\mathcal{Y}_{0:t-1}}{\arg\max} \left(\tilde{f}_{\theta_{t-1}}^{m^\top} \cdot \mu_m\right)==y_m,
\end{equation}
where $\mathbf{idx}$ denotes a boolean tensor that selects the synthetic features $\tilde{f}_{\theta_{t-1}}^{m}\left[\mathbf{idx}\right]$ and $\tilde{f}_{\theta_{t}}^{m}\left[\mathbf{idx}\right]$ whose labels match those of the prototypes. Ultimately, these retrieved synthetic features are exploited to enable effective IPAD, formulated as follows:
\begin{equation}
    \mathcal{L}_{AD}=\mathbb{E}_{(x_t,y_t)\in\mathcal{S}_t}\left[\sum_{m\in\mathcal{Y}_{0:t-1}}\|\tilde{f}_{\theta_{t}}^{m}\left[\mathbf{idx}\right]^{\top}\cdot\mu_{m}-\tilde{f}_{\theta_{t-1}}^{m}\left[\mathbf{idx}\right]^{\top}\cdot\mu_{m}\|_2\right],
\end{equation}
where $\mathcal{S}_t \subseteq \mathcal{T}_t$, subset sampling, particularly on large-scale datasets, optimizes computational efficiency while regularizing model parameters to resist feature drift and catastrophic forgetting.

\subsection{Decision Boundary Perception}

PCL capitalizes on instance-prototype relationships for incremental learning, minimizing the distance between instances and prototypes with congruent labels, while maximizing the separation between those with disparate labels. Prototypes are typically assumed to be sufficiently representative class-level features. However, in practice, class distributions may exhibit considerable diversity, rendering a single prototype inadequate for comprehensive representation. When generalized to new tasks, this can give rise to distributional entanglement across classes, resulting in inter-class ambiguity.

In this paper, we augment PCL with a DBP mechanism, which facilitates explicit inter-class disentanglement by capturing relational dynamics with boundary-adjacent instances. Information entropy~\cite{shannon1953lattice, shannon1948mathematical} is harnessed to quantify the inherent predictive uncertainty of each node. High-entropy nodes, often residing near decision boundaries as hard instances, can be leveraged to promote sharper inter-class separation. Given the embedding $f_{\theta_t}^c$ of a node from task $\mathcal{T}_t$ with class $c$, its information entropy is calculated as follows:
\begin{equation}
    p(x_t^c)=softmax(f_{\theta_{t}}^{c\top}\cdot\left[\mu_{k}\right]_{k\in \mathcal{Y}_{0:t}}),\quad\mathcal{H}(x_t^c)=-\sum_{k\in\mathcal{Y}_{0:t}}p(x_t^c)\log p(x_t^c),
\end{equation}
where $\left[\mu_{k}\right]_{k\in \mathcal{Y}_{0:t}}$ represents concatenated offline and online prototypes. Next, the Top-$K$ highest-entropy instances from each class in task $\mathcal{T}_t$ are identified as hard examples and seamlessly incorporated into PCL training. They dynamically sustain class boundaries and, in conjunction with prototype representations, foster explicit demarcation between both novel and previously learned classes. Eq. \ref{eq2} is further reformulated as:
\begin{equation}\label{eq10}
    \mathcal{L}'_{PCL}=\mathbb{E}_{(x_t,y_t)\in\mathcal{T}_t}\left[\!-\log\frac{e^{\mathcal{F}_{\theta_t}(x_t^c)^{\top}\cdot \mu_c/\tau}}{e^{\mathcal{F}_{\theta_t}(x_t^c)^{\top}\cdot \mu_c/\tau}+\sum_{j\neq c}e^{\mathcal{F}_{\theta_t}(x_t^c)^{\top}\cdot \mu_j/\tau}+\sum_{c'\neq c}e^{\mathcal{F}_{\theta_t}(x_t^c)^{\top}\cdot \mathcal{F}_{\theta_t}(x_t^{c'})/\tau}}\!\right]\!,
\end{equation}
where $c'\in\mathcal{Y}_{t}$, and $x_t^{c'}$ is drawn from the retrieved hard examples. Moreover, akin to conventional PR, $K$ historical embeddings are stochastically sampled from $\mathcal{N}(\mu_m,\sigma_m^2)$ for each previously observed class $m$, and paired with the current instance $x_t^c$ to constitute negative pairs for PCL. Owing to space limitations, this term is omitted from the denominator in Eq.~\ref{eq10}.

\subsection{Feature Drift Compensation}

While IPAD and PCL exhibit notable efficacy in suppressing feature drift, we further rectify the retained prototypes in the memory buffer $\mathcal{M}$ via post-task drift compensation informed by the current task data $(x_t,y_t)\in\mathcal{T}_t$. The calculation procedure is as follows:
\begin{equation}\label{eq_fdc}
    \mu'_{m}=\mu_m+\beta\Delta\mu_m,\forall m\in\mathcal{Y}_{0:t-1},
\end{equation}
where $\Delta\mu_m=\sum_{(x_t,y_t)\in\mathcal{T}_t} w(x_t,\mu_m)(\mathcal{F}_{\theta_{t}}(x_t)-\mathcal{F}_{\theta_{t-1}}(x_t))$, and $\beta$ is a hyperparameter controlling compensation intensity. $w(x_t,\mu_m)$ quantifies the proximity between node $x_t$ and prototype $\mu_m$, with closer nodes contributing more to drift compensation. The formal definition is given as follows:
\begin{equation}
    w(x_t,\mu_m)=\frac{\mathcal{F}_{\theta_{t-1}}(x_t)^{\top}\cdot\mu_m}{\sum_{(x'_t,y'_t)\in\mathcal{T}_t}\mathcal{F}_{\theta_{t-1}}(x'_t)^{\top}\cdot\mu_m}.
\end{equation}
In this paper, in light of the observation in Figure~\ref{fig1}, $\beta$ is assigned an exceedingly small value.

\subsection{Optimization Objective}

To train the proposed IPAL, the overall optimization objective is as follows:
\begin{equation}\label{eq11}
    \mathcal{L}=\mathcal{L}'_{PCL}+\gamma\mathcal{L}_{AD},
\end{equation}
where $\gamma$ denotes the weighting factor, governing the trade-off between plasticity and stability. Notably, for the base task $\mathcal{T}_0$, the optimization objective is confined to $\mathcal{L}'_{PCL}$, with $\gamma=0$. Algorithm~\ref{alg} provides a detailed depiction of the training workflow.

\begin{algorithm}[t]
\caption{Training procedure for our IPAL.}
\label{alg}
\KwIn{Task sequence $\mathcal{T}=\{\mathcal{T}_{0}, \mathcal{T}_{1},...,\mathcal{T}_{N}\}$, GNN encoder $\mathcal{F}_{\theta}(\cdot)$, memory buffer $\mathcal{M}$, weighting factor $\alpha$, base task learning rate $\eta_0$, incremental task learning rate $\eta_{t>0}$, number of epochs $E$.}
\KwOut{Predicted labels for test nodes from all previously learned tasks.}
\For{$t=0,1,...,N$}{
    \For{$epoch=1,2,...,E$}{
        \If{$t=0$}{\tcp{Initial training on $\mathcal{T}_0$.}
            Compute $\mathcal{L}'_{PCL}$ according to Eq. \ref{eq10}, i.e., Eq. \ref{eq11} with $\gamma=0$.\\
            $\theta_t \leftarrow\theta_t-\eta_0\nabla_{\theta_t}\mathcal{L}'_{PCL}$.\\
        }
        \Else{\tcp{Incremental training on $\mathcal{T}_{t>0}$.}
            Compute $\mathcal{L}$ according to Eq. \ref{eq11}.\\
            $\theta_t \leftarrow\theta_t-\eta_t\nabla_{\theta_t}\mathcal{L}$.\\
        }
    }
    Perform drift compensation on prior class prototypes $\{\mu_m\}_{m\in\mathcal{Y}_{0:t-1}}$ according to Eq. \ref{eq_fdc}.\\
    Generate TIGP $\{\mathcal{N}(\mu_c,\sigma_{c}^2)\}_{c\in\mathcal{Y}_{t}}$ according to Eq. \ref{eq4}, and allocate them to the memory buffer $\mathcal{M}$.\\
    \tcp{Testing phase.}
    Predict node labels for all prior task graphs: $\hat{y}(x)=\underset{k\in\mathcal{Y}_{0:t}}{\arg\max} \mathcal{F}_{\theta_t}(x)^{\top}\cdot\left[\mu_{k}\right], \forall x\in\mathcal{T}_{0:t}$.
}
\Return $\hat{y}(x)$.
\end{algorithm}

\section{Experiments}
\label{exp}

In this section, we empirically investigate the following questions: \textbf{Q1}) Does IPAL yield performance gains over existing state-of-the-art NECL methods? \textbf{Q2}) Do the proposed components substantively bolster the overall effectiveness of IPAL? \textbf{Q3}) How does the weighting factor $\gamma$ modulate the performance of IPAL?

\subsection{Experimental Setup}
\label{exp1}

\textbf{Datasets.} We evaluate IPAL on four node classification benchmark datasets: CS-CL~\cite{shchur2018pitfalls}, CoraFull-CL~\cite{mccallum2000automating}, Arxiv-CL~\cite{hu2020open} and Reddit-CL~\cite{hamilton2017inductive}. Following the problem formulation, each dataset is partitioned into a base task $\mathcal{T}_0$ and a series of incremental tasks $\{\mathcal{T}_{t}\}_{t=1}^{t=N}$ via label-wise stratification. The graph is decomposed into $N+1$ disjoint subgraphs, each dedicated to a specific training task. All tasks follow a class-wise 6/2/2 split for training/validation/testing. The suffix "-CL" signifies the generated task sequence for CGL. The comprehensive descriptions are delineated in Appendix~\ref{appendix:b1}.

\textbf{Implementation Details.} IPAL takes a 2-layer GCN~\cite{kipf2016semi} with a hidden dimension of 128, initializing the learning rate with $\eta_0=1\times 10^{-3}$ for the base task and $\eta_{t>0}=1\times10^{-4}$ for incremental tasks. The algorithm implementation and task training are grounded in Continual Graph Learning Benchmark (CGLB)~\cite{zhang2022cglb}, with all experiments executed within the PyTorch 3.10 framework powered by an NVIDIA 3090 GPU. We set the temperature scaling factor $\tau=0.07$, damping factor $\alpha=0.85$, compensation intensity $\beta=0.1$, $\left|\mathcal{S}_t\right|=100$, and $K=10$. The weighting factor $\gamma$ is tuned via grid search over $\left[0.1,1.0\right]$ with a step size of 0.1. Full-graph training is executed on CS-CL and CoraFull-CL, while mini-batch training, with a batch size of 2000, is applied to the larger-scale Arxiv-CL and Reddit-CL datasets. All experiments are run 5 times, with the mean and standard deviation reported.

\textbf{Baselines and Evaluation Metrics.} We compare our IPAL with existing state-of-the-art methods, including regularization-based methods (i.e., EWC~\cite{kirkpatrick2017overcoming}, MAS~\cite{aljundi2018memory}, LWF~\cite{li2017learning}, GEM~\cite{lopez2017gradient}, and TWP~\cite{liu2021overcoming}), Non-Exemplar methods (i.e., POLO~\cite{wang2023non}, and EFC~\cite{magistri2024elastic}), and a classic rehearsal-based method, ER-GNN~\cite{zhou2021overcoming}. Furthermore, we consider two canonical baselines: Bare, naive fine-tuning without any auxiliary strategy, and Joint, ideal joint training across tasks, serving as the empirical lower and upper bounds. Average Performance (AP) and Average Forgetting (AF) evaluate the overall classification efficacy and cumulative forgetting across all prior tasks. If a model excels in both metrics, it indicates a balanced trade-off between plasticity and stability. The mathematical definitions are elucidated in Appendix~\ref{appendix:b2}.

\subsection{Q1: Comparison with the State-of-the-Art}

\begin{table}[t]
\centering
\caption{Performance comparison with existing state-of-the-art baselines on CS-CL, CoraFull-CL, Arxiv-CL, and Reddit-CL. The best results are highlighted in bold, and the second-best results are underlined.}
\label{tab1}
\resizebox{\textwidth}{!}{
\begin{tabular}{c||cc|cc|cc|cc}
\toprule
                          & \multicolumn{2}{c|}{CS-CL}                                      & \multicolumn{2}{c|}{CoraFull-CL}                                & \multicolumn{2}{c|}{Arxiv-CL}                                   & \multicolumn{2}{c}{Reddit-CL}                                  \\ \cline{2-9} 
\multirow{-2}{*}{Methods} & AP/\% $\uparrow$                              & AF/\% $\uparrow$                     & AP/\% $\uparrow$                              & AF/\% $\uparrow$                    & AP/\% $\uparrow$                             & AF/\% $\uparrow$                    & AP/\% $\uparrow$                             & AF/\% $\uparrow$                     \\ \midrule\midrule
Joint                     & 95.53$\pm$0.10 & - & 74.66$\pm$0.47 & - & 55.19$\pm$0.76 & - & 96.11$\pm$0.12 & - \\
Bare                      & 43.06$\pm$3.95                         & -63.68$\pm$4.84               & 14.63$\pm$0.48                         & -75.57$\pm$0.58               & 16.30$\pm$0.18                         & -74.41$\pm$0.38               & 19.47$\pm$0.58                         & -96.42$\pm$0.69               \\ \midrule
EWC                       & 48.11$\pm$4.73                         & -57.49$\pm$5.84               & 15.32$\pm$0.95                         & -71.17$\pm$2.15               & 18.04$\pm$1.26                         & -67.91$\pm$2.50               & 20.77$\pm$0.89                         & -94.73$\pm$1.25               \\
MAS                       & 53.97$\pm$3.67                         & -48.25$\pm$4.47               & 16.82$\pm$0.77                         & -65.27$\pm$1.19               & 21.66$\pm$2.93                         & -45.83$\pm$6.20               & 20.03$\pm$1.46                         & -94.13$\pm$1.66               \\
LWF                       & 52.41$\pm$3.04                         & -52.45$\pm$3.67               & 16.51$\pm$0.84                         & -69.31$\pm$1.35               & 16.48$\pm$0.33                         & -74.73$\pm$0.18               & 22.59$\pm$2.42                         & -92.42$\pm$2.94               \\
GEM                       & 56.56$\pm$5.06                         & -47.44$\pm$6.15               & 19.62$\pm$0.23                         & -64.62$\pm$0.68               & 19.36$\pm$0.48                         & -68.06$\pm$0.59               & 44.99$\pm$5.78                         & -64.04$\pm$7.37               \\
TWP                       & 55.98$\pm$4.17                         & -47.98$\pm$5.29               & 15.65$\pm$0.89                         & -73.85$\pm$1.82               & 18.58$\pm$2.16                         & -67.38$\pm$4.04               & 23.23$\pm$3.21                         & -91.65$\pm$4.02               \\
ER-GNN                    & \underline{81.19$\pm$1.82}                   & -17.15$\pm$2.20               & 19.54$\pm$0.43                         & -67.41$\pm$0.54               & 27.13$\pm$0.34                         & -56.55$\pm$0.43               & 84.64$\pm$2.04                         & -14.06$\pm$2.53               \\
POLO                      & 60.68$\pm$2.32                         & -16.96$\pm$3.90               & 23.27$\pm$1.60                         & \textbf{-18.03$\pm$1.92}      & \underline{32.03$\pm$0.75}                   & -45.48$\pm$0.86               & \underline{92.07$\pm$1.13}                   & \underline{-1.76$\pm$0.79}          \\
EFC                       & 71.65$\pm$2.51                         & \underline{-16.35$\pm$3.24}         & \underline{38.75$\pm$1.85}                   & \underline{-19.55$\pm$2.02}         & 30.94$\pm$0.41                         & \underline{-40.13$\pm$1.48}         & 87.49$\pm$1.92                         & -8.20$\pm$2.03                \\ \midrule
IPAL                      & \textbf{83.07$\pm$2.16}                & \textbf{-12.89$\pm$2.50}      & \textbf{40.69$\pm$2.54}                & -20.60$\pm$0.68               & \textbf{33.10$\pm$0.84}                & \textbf{-20.59$\pm$0.69}      & \textbf{92.15$\pm$0.13}                & \textbf{-0.27$\pm$0.19}       \\ \bottomrule
\end{tabular}
}
\end{table}

\begin{figure}[t]
    \centering
    \includegraphics[width=\linewidth]{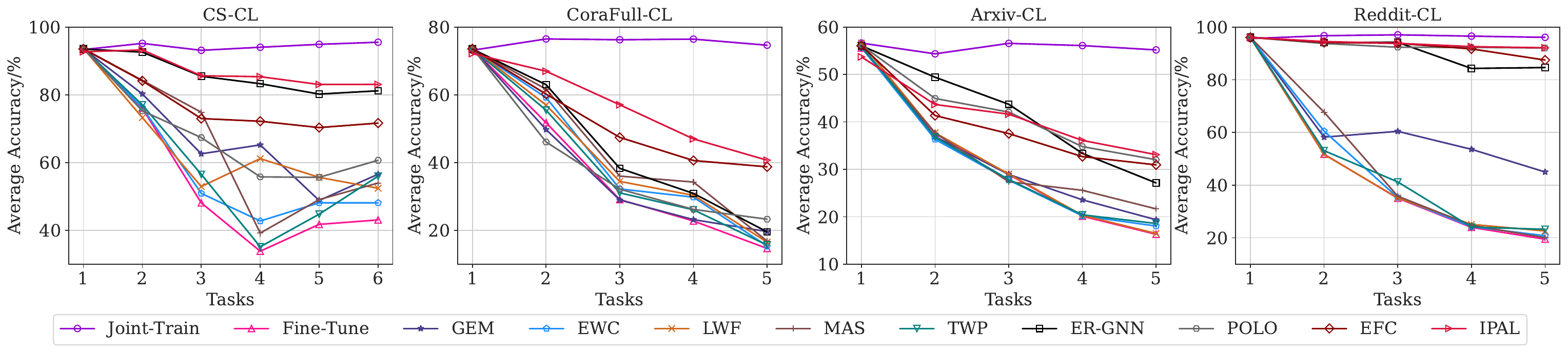}
    \caption{Learning dynamics over the task sequences on CS-CL, CoraFull-CL, Arxiv-CL, and Reddit-CL. The AP is reported on all tasks.}
    \label{fig4}
\end{figure}

As shown in Table~\ref{tab1}, we compare our IPAL with several state-of-the-art methods on four node classification benchmark datasets. Without intervention for catastrophic forgetting, Bare performs direct fine-tuning on incremental tasks, leading to considerable erosion of prior knowledge. While regularization-based methods mitigate catastrophic forgetting to some extent, their efficacy falls short of rehearsal-based and Non-Exemplar approaches. TWP, tailored for graph-structured tasks, yet in this context, performs comparably to traditional methods. ER-GNN revisits pivotal historical examples to retain previous memory, yet it considers only individual nodes, disregarding the significance of topological structure. In contrast, Non-Exemplar methods replay historical prototypes that encapsulate both class-wise features and topological information, outperforming ER-GNN on CoraFull-CL, Arxiv-CL, and Reddit-CL. However, due to the inherent flaw of conventional PR, feature drift progressively exacerbates with continual model updates. Our proposed IPAL capitalizes on the PCL paradigm to curb feature drift, consistently outstripping existing methods on four benchmark datasets. Notwithstanding a minor decline in AF on CoraFull-CL, IPAL achieves pronounced gains in AP. This discrepancy stems from the inclination of POLO and EFC toward favoring stability over plasticity, whereas our IPAL achieves a more balanced trade-off, yielding superior performance overall. Figure~\ref{fig4} illustrates the learning dynamics on four datasets, highlighting that our IPAL almost invariably outperforms the competing approaches throughout the entire task sequences.

\subsection{Q2: Ablation Studies and Visualization}

\begin{figure}[t]
\begin{minipage}{0.65\textwidth}
\centering
\includegraphics[width=0.5\linewidth, trim=20 20 20 20, clip]{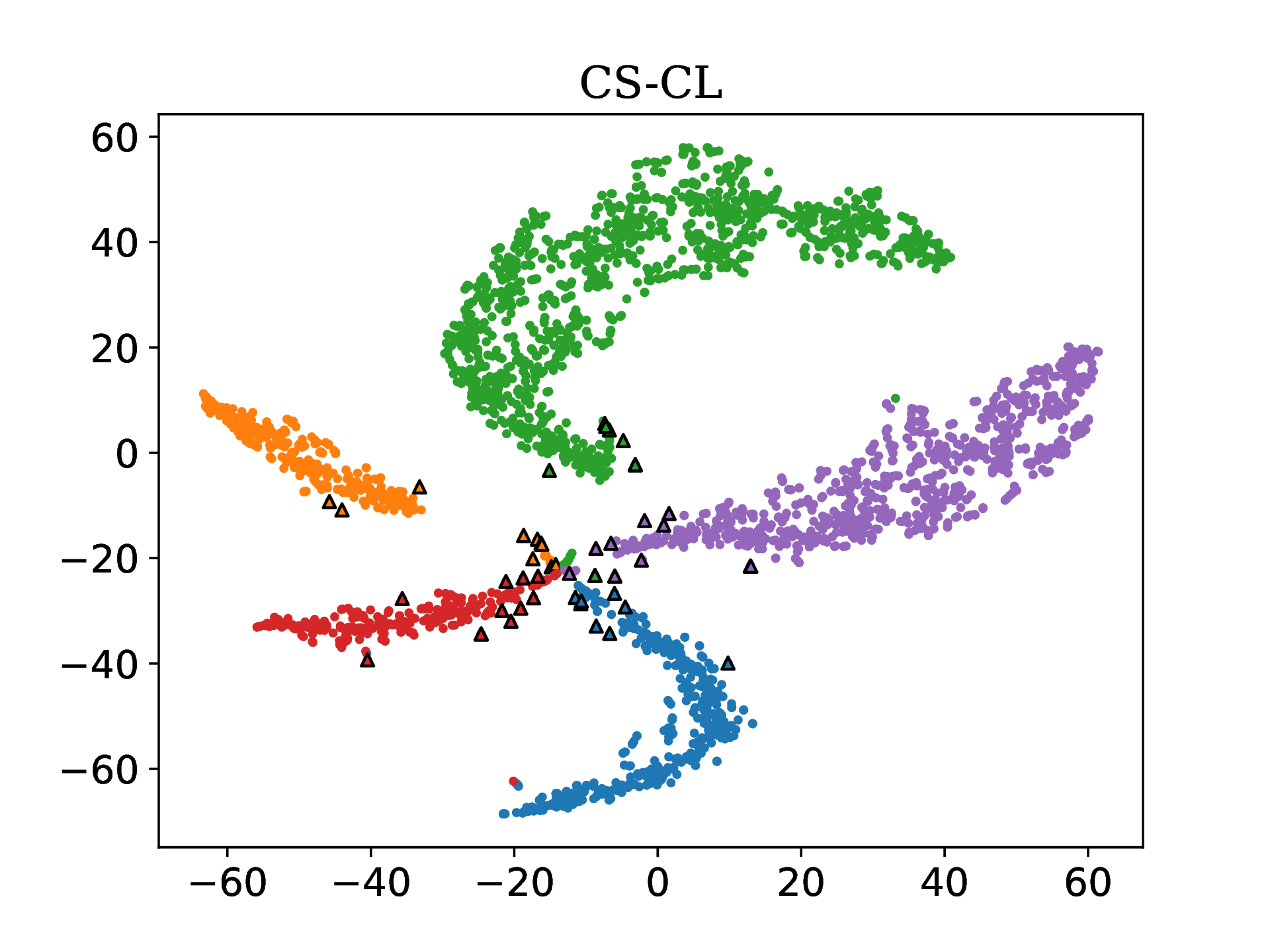}
\hspace{-10pt}
\includegraphics[width=0.5\linewidth, trim=20 20 20 20, clip]{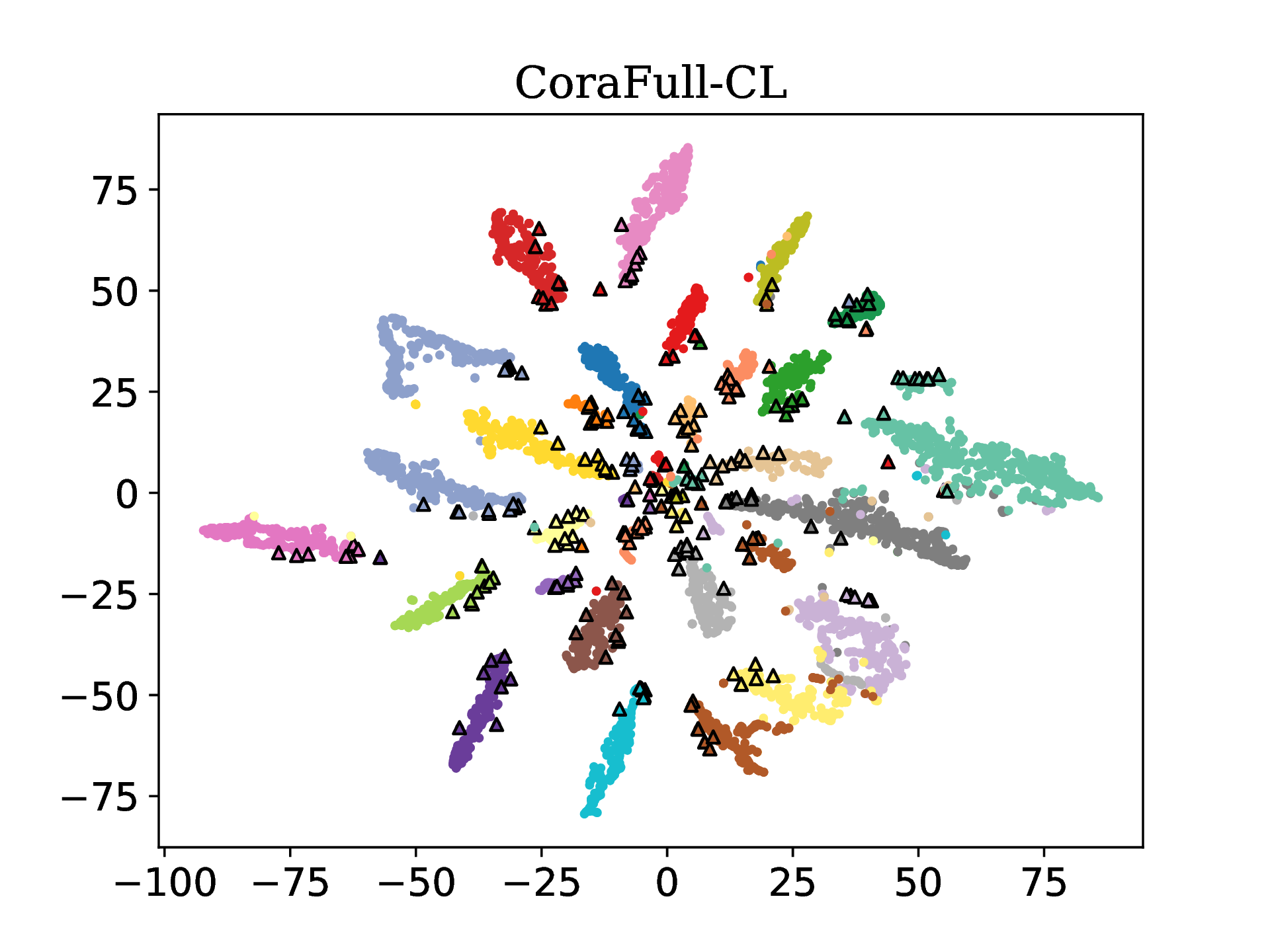}
\caption{Visualization of class distributions on the base task $\mathcal{T}_0$. Each color corresponds to a specific class, and triangles indicate hard examples.}
\label{fig5}
\end{minipage}
\hfill
\begin{minipage}{0.33\textwidth}
\vspace{11pt}
\includegraphics[width=0.95\linewidth]{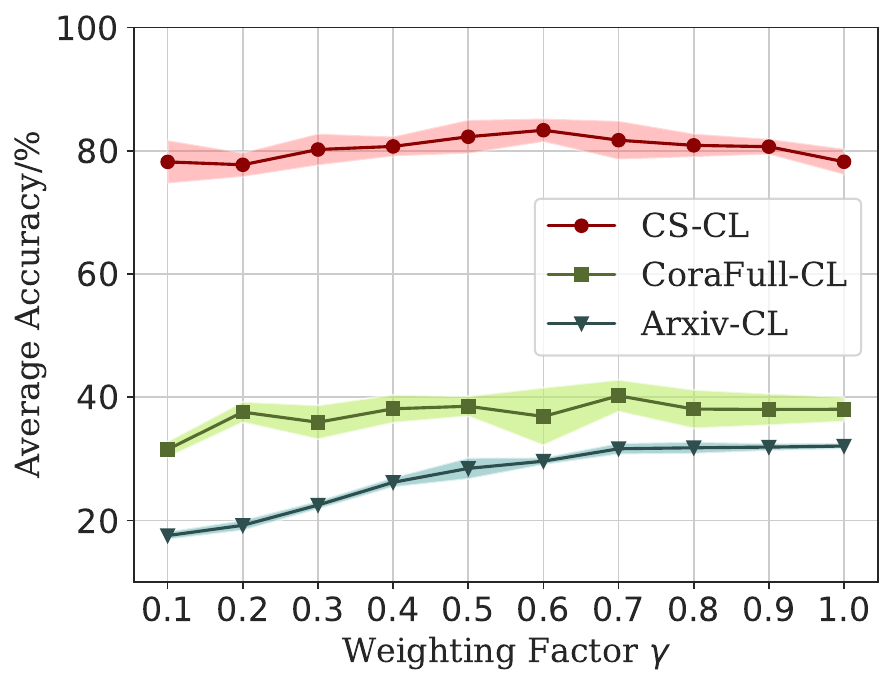}
\vspace{-4pt}
\caption{Comparison of various $\gamma$ settings on CS-CL, CoraFull-CL, and Arxiv-CL, with AP reported.}
\label{fig6}
\end{minipage}
\end{figure}

To further validate the proposed components, we conduct ablation studies on CS-CL, CoraFull-CL, and Arxiv-CL. Table~\ref{tab2} reveals the following observations: \romannumeral1) Replacing the topology-aware TIGP with simplistic mean-based prototypes incurs a substantial performance decline, most notably on CS-CL and Arxiv-CL; \romannumeral2) Eliminating distillation (including IPAD and FD) leads to severe catastrophic forgetting, yielding a performance decline exceeding 8\% on each dataset; 
\begin{wraptable}{r}{0.6\textwidth}
\centering
\caption{Ablation studies on CS-CL, CoraFull-CL, and Arxiv-CL. The AP is reported, with the best results highlighted in bold. FD refers to Feature Distillation.}
\label{tab2}
\resizebox{0.6\textwidth}{!}{
\begin{tabular}{cccc||ccc}
\toprule
TIGP & IPAD & FD & DBP & CS-CL      & CoraFull-CL & Arxiv-CL   \\ \midrule\midrule
\XSolidBrush & \Checkmark & \XSolidBrush  & \Checkmark    & 79.93$\pm$1.85 & 39.98$\pm$1.44  & 30.42$\pm$0.74 \\
\Checkmark     & \XSolidBrush    & \XSolidBrush  &  \Checkmark   & 64.07$\pm$4.32 & 32.50$\pm$2.76  & 16.89$\pm$0.25 \\
  \Checkmark   & \XSolidBrush    &  \Checkmark  &   \Checkmark  & 72.79$\pm$2.70 & 36.04$\pm$2.83  & 27.25$\pm$0.71 \\
  \Checkmark   &  \Checkmark    & \XSolidBrush  & \XSolidBrush   & 75.94$\pm$3.05 & 33.63$\pm$2.58  & 31.06$\pm$0.42 \\
  \Checkmark   &   \Checkmark   & \XSolidBrush  &   \Checkmark  & \textbf{83.07$\pm$2.16} & \textbf{40.69$\pm$2.54}  & \textbf{33.10$\pm$0.84} \\ \bottomrule
\end{tabular}
}
\vspace{-10pt}
\end{wraptable}
\romannumeral3) While FD proves effective in mitigating catastrophic forgetting and feature drift, it may inadvertently impose excessive constraints, hindering the model's capacity to assimilate novel knowledge; \romannumeral4) DBP takes into account decision-boundary instances in the PCL objective, further promoting inter-class separation. Moreover, we provide visualizations to elucidate the contributions of the proposed components in a more intuitive manner. Refer to Section~\ref{Methodology} for the analysis of Figure~\ref{fig3}. Figure~\ref{fig5} visualizes the class distributions of the base task $\mathcal{T}_0$ on CS-CL and CoraFull-CL, with DBP-identified hard examples located at cluster boundaries, validating our previous analysis.

\subsection{Q3: Parameter Analysis}
\label{exp3}

A grid search over $\left[0.1,1.0\right]$ is carried out on the validation sets of CS-CL, CoraFull-CL, and Arxiv-CL to examine performance sensitivity to the weighting factor $\gamma$. As shown in Figure~\ref{fig6}, the optimal performance is attained with a weighting factor of 0.6 or 0.7 on small-scale datasets, whereas larger datasets require a higher weight. Overall, smaller $\gamma$ hinders task retention (i.e., stability $\downarrow$, plasticity $\uparrow$), while larger values impose undue constraints (i.e., stability $\uparrow$, plasticity $\downarrow$). Empirically, a value around 0.7 strikes a favorable trade-off between plasticity and stability, consistently yielding superior performance on all datasets and streamlining the hyperparameter tuning process.

\section{Conclusions and Future Works}

In this paper, we empirically revealed that PCL substantially attenuates feature drift compared to conventional PR by harnessing the intrinsic relational topology between instances and prototypes. We further proposed IPAL, a novel NECGL paradigm built upon PCL. To be specific, we evaluated node importance via the PageRank algorithm and generated the topology-aware TIGP to promote PCL training. To address the inherent feature drift and catastrophic forgetting in NECGL, IPAD was proposed to regularize the relational structure between instances and prototypes, providing greater flexibility while seamlessly aligning with the PCL objective function. Furthermore, the DBP mechanism was leveraged to mine hard examples, mitigating inter-class ambiguity and fostering more pronounced inter-class separability. Extensive empirical evaluations on four node classification benchmark datasets, including comparative, ablation, and parameter studies, demonstrated that IPAL consistently outperformed existing state-of-the-art methods. Our future work will seek to adapt IPAL to online settings with streaming data, promoting its scalability and real-world viability.



\bibliographystyle{plain}
\bibliography{ref}


\appendix

\section{Theoretical Analysis on Feature Drift}
\label{appendix:a}

\textbf{Notation.} Let $\mathcal{F}_{\theta}(\cdot): x\mapsto f$ denote the GNN encoder, inherently Lipschitz continuous with respect to the input~\cite{ruiz2021graph}. Access pertaining to task $\mathcal{T}_t$ is confined to the current task data $(x_t,y_t)\in\mathcal{T}_t$ and the prototypes $(\mu_m,\Sigma_m)$ stored in the memory buffer $\mathcal{M}$, where $m\in\mathcal{Y}_{0:t-1}$. After training on task $\mathcal{T}_t$, the GNN encoder is updated from $\theta_{t-1}$ to $\theta_t$. Consider the samples $(x_m,y_m)\in\mathcal{T}_{0:t-1}$ from previous tasks. Let the feature distributions follow a Gaussian distribution, where $\mathcal{F}_{\theta_t}(x_m)\sim \mathcal{P}_t(f)=\mathcal{N}(\mu'_m,\Sigma'_m)$ and $\mathcal{F}_{\theta_{t-1}}(x_{m})\sim \mathcal{P}_{t-1}(f)=\mathcal{N}(\mu_{m},\Sigma_{m})$. Note that the notation here differs slightly from that in the main text, such as the covariance $\Sigma_m$ and the diagonal covariance $\sigma_m^2$.

In Non-Exemplar Continual Learning (NECL), Prototype Contrastive Learning (PCL) induces less drift than conventional Prototype Replay (PR). To validate this theoretically, we adopt the Kullback-Leibler (KL) divergence to quantify feature drift and investigate the evolution of feature distributions from prior tasks during the assimilation of new tasks for both PCL and PR, as follows:
\begin{equation}
    D_{KL}(\mathcal{P}_{t-1}\|\mathcal{P}_t)=\int \mathcal{P}_{t-1}(f)\log\frac{\mathcal{P}_{t-1}(f)}{\mathcal{P}_t(f)}\,df.
\end{equation}
The goal is to prove that the KL divergence for PCL is theoretically less than that for PR, as formulated below:
\begin{theorem}
\label{theo1}
     In NECL, assuming Gaussian feature distributions with $\mathcal{F}_{\theta_t}(x_m)\sim \mathcal{P}_t(f)=\mathcal{N}(\mu'_m,\Sigma'_m)$ and $\mathcal{F}_{\theta_{t-1}}(x_{m})\sim \mathcal{P}_{t-1}(f)=\mathcal{N}(\mu_{m},\Sigma_{m})$, where $\Sigma'_m$ and $\Sigma_{m}$ are positive definite, PCL incurs a smaller feature drift than PR:
     \begin{equation}
         D_{KL}(\mathcal{P}_{t-1}\|\mathcal{P}_{t})_{PCL}<D_{KL}(\mathcal{P}_{t-1}\|\mathcal{P}_{t})_{PR}.
     \end{equation}
\end{theorem}

\textbf{Proof.} For the $n$-dimensional multivariate Gaussian distribution, the expression has a closed-form solution as follows:
\begin{equation}
    D_{KL}(\mathcal{P}_{t-1}\|\mathcal{P}_t)=\frac{1}{2}\left(\log\frac{\det \Sigma'_{m}}{\det \Sigma_{m}}+\text{tr}(\Sigma_m^{'-1}\Sigma_{m})+(\mu'_{m}-\mu_{m})^{\top}\Sigma_m^{'-1}(\mu'_{m}-\mu_{m})-n\right).
\end{equation}
To prove Theorem~\ref{theo1}, it suffices to show that for prior feature distributions, PCL yields smaller $\|\mu'_{m}-\mu_{m}\|_2^2$ and $\text{tr}(\Sigma_m^{'-1}\Sigma_{m}-\text{I})$ than PR.

PCL updates $\theta_t$ on task $\mathcal{T}_t$ by optimizing Eq.~\ref{eq2}, with the feature gradient computed as follows:
\begin{equation}\label{eq17}
    \nabla_{f_{\theta_t}}\mathcal{L}_{PCL}\propto -\frac{1}{\tau}\left[\mu_{c}-\sum_{k\in\mathcal{Y}_{0:t}}\frac{e^{f_{\theta_t}^{\top}\cdot\mu_{k}/\tau}}{\sum_{k'\in\mathcal{Y}_{0:t}}e^{f_{\theta_t}^{\top}\cdot\mu_{k'}/\tau}}\mu_{k}\right].
\end{equation}
The gradient for the model parameters $\theta_t$ is computed as follows:
\begin{equation}
\nabla_{\theta}\mathcal{L}_{PCL}=\nabla_{f_{\theta_t}}\mathcal{L}_{PCL}\cdot\nabla_{\theta}\mathcal{F}_{\theta_{t}}(x_t).
\end{equation}
Considering $(x_m,y_m)\in\mathcal{T}_{0:t-1}$, we have:
\begin{equation}
    \mathcal{F}_{\theta_{t}}(x_m)=\mathcal{F}_{\theta_{t-1}+\Delta\theta}(x_m).
\end{equation}
Here, considering an infinitesimal update step $\Delta\theta$, it can be approximated by the first-order Taylor expansion~\cite{zenke2017continual} as follows:
\begin{equation}
    \mathcal{F}_{\theta_t}(x_m)\approx\mathcal{F}_{\theta_{t-1}}(x_m)+\nabla_\theta\mathcal{F}_{\theta_{t-1}}(x_m)\cdot\Delta\theta.
\end{equation}
Furthermore, we obtain:
\begin{equation}
    \mu'_m\approx\mu_{m}+\mathbb{E}_{(x_m,y_m)\in\mathcal{T}_{0:t-1}}\left[\nabla_{\theta}\mathcal{F}_{\theta_{t-1}}(x_m)\cdot\Delta\theta\right].
\end{equation}
Eq.~\ref{eq2} regularizes the negative sample gradients in Eq.~\ref{eq17} by minimizing $\mathcal{F}_{\theta_t}(x_t)^\top\cdot\mu_m$, encouraging $\Delta\theta$ to drive $\mathcal{F}_{\theta_t}(x_m)$ closer to $\mathcal{F}_{\theta_{t-1}}(x_m)$. Thus, $\mathcal{L}_{PCL}$ essentially regularizes $\|\mu'_{m}-\mu_{m}\|_2^2$, effectively alleviating feature shift.

PR optimizes Eq.~\ref{eq1} during training, with the feature gradient computed as follows:
\begin{equation}
    \nabla_{f_{\theta_{t}}}\mathcal{L}_{PR}\propto\sum_{k\in\mathcal{Y}_{0:t}}\left(\frac{e^{f_{\theta_t}^\top\cdot w_k}}{\sum_{k'\in\mathcal{Y}_{0:t}}e^{f_{\theta_t}^\top\cdot w_{k'}}}-\mathds{1}\{k=y_t^c\}\right)w_{k},
\end{equation}
where $w_k$ denotes the class-$k$ weight of the linear classifier $g_{\phi_t}(\cdot)$. Since PR exclusively emphasizes the classification accuracy of the new sample $(x_t,y_t)\in\mathcal{T}_t$ without explicitly regularizing $\Delta\theta$ to maintain the alignment between $\mathcal{F}_{\theta_t}(x_m)$ and $\mathcal{F}_{\theta_{t-1}}(x_m)$, it may exacerbate the deviation, leading to an increase in $\|\mu'_{m}-\mu_{m}\|_2^2$.

On the other hand, the covariance of past tasks is defined as:
\begin{equation}
    \Sigma'_m=\mathbb{E}_{(x_m,y_m)\in\mathcal{T}_{0:t-1}}\left[\left(\mathcal{F}_{\theta_{t}}(x_m)-\mu'_m\right)\left(\mathcal{F}_{\theta_{t}}(x_m)-\mu'_m\right)^\top\right].
\end{equation}
According to our preceding analysis, PCL encourages alignment between $\mathcal{F}_{\theta_{t}}(x_m)$ and $\mathcal{F}_{\theta_{t-1}}(x_m)$, whereas PR, lacking such a constraint, allows $\Sigma_m$ to expand uncontrollably.

Therefore, we conclude that PCL yields smaller $\|\mu'_{m}-\mu_{m}\|_2^2$ and $\text{tr}(\Sigma_m^{'-1}\Sigma_{m}-\text{I})$ than PR, thereby substantiating the validity of Theorem \ref{theo1}.

\section{Extended Details on Experimental Configuration}

\subsection{Additional Descriptions on the Datasets}
\label{appendix:b1}

Four node classification benchmark datasets are engaged in this paper, with the following detailed descriptions:

\textbf{CS-CL~\cite{shchur2018pitfalls}.} Coauthor CS is a co-authorship graph based on the Microsoft Academic Graph from the KDD Cup 2016 challenge. Nodes stand for authors, linked by edges in cases of co-authorship. Node attributes encode paper keywords, and class labels signify the author's principal research domains. In this study, we divide the dataset into a base task with 5 classes and 5 incremental tasks, each containing 2 of the remaining 10 classes.

\textbf{CoraFull-CL~\cite{mccallum2000automating}.} CoraFull is a more complete citation network dataset than the commonly used 7-class subset, with nodes as papers, labels as topics, and edges as citation links. All papers are classified into 70 discrete topics, from which 30 are designated to constitute the base task, while the remaining 40 are evenly partitioned into 4 incremental tasks of 10 topics each.

\textbf{Arxiv-CL~\cite{hu2020open}.} OGB-Arxiv is a paper citation network of Arxiv papers extracted from the Microsoft Academic Graph. Each node is an Arxiv paper, with directed edges indicating citations between papers. The skip-gram model~\cite{mikolov2013distributed} is applied to extract word embeddings from titles and abstracts, which are then employed to define node attributes. The dataset comprises 40 subject areas from Arxiv Computer Science papers. The first 20 areas form the base task, and the remaining 20 areas are grouped into 4 incremental tasks of 5 areas each.

\textbf{Reddit-CL~\cite{hamilton2017inductive}.} Reddit comprises posts made in September 2014, with each node labeled by its associated subreddit. 41 large communities are taken into account to construct a post-to-post graph, with edges defined by user comments on both posts. Node attributes include the post title, average comment embedding, post score, and comment count. Following \cite{zhang2022cglb}, we exclude the 41$st$ class, then treat the first 20 as the base task and group the remaining 20 into 4 incremental tasks of 5 classes each.

Table~\ref{tab:appendix1} presents the statistical information of the four benchmark datasets.

\begin{table}[t]
\centering
\caption{The statistical information of CS-CL, CoraFull-CL, Arxiv-CL, and Reddit-CL.}
\label{tab:appendix1}
\begin{tabular}{l||clclclcl}
\toprule
\multicolumn{1}{c||}{Benchmark Datasets} & \multicolumn{2}{c}{CS-CL}  & \multicolumn{2}{c}{CoraFull-CL} & \multicolumn{2}{c}{Arxiv-CL} & \multicolumn{2}{c}{Reddit-CL} \\ \midrule\midrule
\# nodes                                & \multicolumn{2}{c}{18333}  & \multicolumn{2}{c}{19793}       & \multicolumn{2}{c}{169343}   & \multicolumn{2}{c}{232965}    \\
\# edges                                & \multicolumn{2}{c}{163788} & \multicolumn{2}{c}{126842}      & \multicolumn{2}{c}{1166243}  & \multicolumn{2}{c}{114615892} \\
\# features                             & \multicolumn{2}{c}{6805}   & \multicolumn{2}{c}{8710}        & \multicolumn{2}{c}{128}      & \multicolumn{2}{c}{602}       \\
\# labels                               & \multicolumn{2}{c}{15}     & \multicolumn{2}{c}{70}          & \multicolumn{2}{c}{40}       & \multicolumn{2}{c}{40}        \\
\# base classes                         & \multicolumn{2}{c}{5}      & \multicolumn{2}{c}{30}          & \multicolumn{2}{c}{20}       & \multicolumn{2}{c}{20}        \\
\# novel classes                        & \multicolumn{2}{c}{10}     & \multicolumn{2}{c}{40}          & \multicolumn{2}{c}{20}       & \multicolumn{2}{c}{20}        \\
\# split                                & \multicolumn{2}{c}{5$+$5$\times$2}  & \multicolumn{2}{c}{30$+$4$\times$10}     & \multicolumn{2}{c}{20$+$4$\times$5}   & \multicolumn{2}{c}{20$+$4$\times$5}    \\
\# tasks                             & \multicolumn{2}{c}{6}      & \multicolumn{2}{c}{5}           & \multicolumn{2}{c}{5}        & \multicolumn{2}{c}{5}         \\ \bottomrule
\end{tabular}
\end{table}

\subsection{Mathematical Definitions for the Evaluation Metrics}
\label{appendix:b2}

Average Performance (AP) and Average Forgetting (AF) quantify the overall classification performance and the extent of catastrophic forgetting, respectively. Their mathematical definitions are as follows:
\begin{equation}
\label{eq:appendix1}
\mathrm{AP}_t=\frac{\sum_{i=1}^{t}\mathbf{M}_{t,i}}{t},\quad\mathrm{AF}_t=\frac{\sum_{i=1}^{t-1}\mathbf{M}_{t,i}-\mathbf{M}_{i,i}}{t-1},
\end{equation}
where $\mathbf{M}$ denotes the lower triangular performance matrix (refer to Figure~\ref{fig3}). $\mathrm{M}_{t,i}$ is the prediction accuracy on task $\mathcal{T}_i$ after training on task $\mathcal{T}_t$, with $t$ indexed from 1 for convenience. Owing to the inherent trade-off between AP and AF, an elevated AP may compromise AF, and vice versa. Both metrics should be taken into account in performance evaluation.

\section{Limitations}
\label{appendix:c}

While our method exhibits competitive performance in Non-Exemplar Continual Graph Learning, several limitations merit further exploration: \romannumeral1) Existing approaches, including ours, assume a unified dataset where tasks are simulated by class-based subgraph partitioning within a single graph. However, in real-world scenarios involving heterogeneous domains, the cross-domain generalizability remains to be validated; \romannumeral2) Our method is trained offline, but its applicability to online settings, where data arrives in a mini-batch streaming manner, remains to be further investigated.


\end{document}